\documentclass[letterpaper, 10 pt, conference]{ieeeconf}

\usepackage{color}
\usepackage{graphicx}
\usepackage{epsbox}
\usepackage{mathptmx}
\usepackage{amsfonts}

\usepackage{times}
\newcounter{Rulenum}
\definecolor{red}{rgb}{1,0,0}
\definecolor{green}{rgb}{0,1,0}
\definecolor{blue}{rgb}{0,0,1}
\definecolor{violet}{rgb}{1,0,1}
\definecolor{cyan}{cmyk}{1,0,0,0}
\definecolor{magenta}{cmyk}{0,1,0,0}
\definecolor{yellow}{cmyk}{0,0,1,0}

\newcommand{\FIG}[3]{
\begin{minipage}[b]{#1cm}
\begin{center}
\includegraphics[width=#1cm]{#2}
{\scriptsize #3}
\vspace*{1mm}
\end{center}
\end{minipage}
}

\newcommand{\acu}[1]{$\acute{\mbox{#1}}$}

\newcommand{\FIGR}[3]{
\begin{minipage}[b]{#1cm}
\begin{center}
\includegraphics[angle=-90,clip,width=#1cm]{#2}
{\scriptsize #3}
\vspace*{1mm}
\end{center}
\end{minipage}
}

\def\Rm#1{
    \uppercase\expandafter{\romannumeral#1}
}

\usepackage{multirow}
\newcommand{\eq}[1]{\begin{equation}#1\end{equation}}

\begin{document}

\title{\Huge Incremental RANSAC for Online Relocation in Large Dynamic Environments}
\author{\authorblockN{Kanji Tanaka$$~~~~~~~~Eiji Kondo$$}
\authorblockA{Graduate School of Engineering\\
Kyushu University\\
744 Motooka, Nishi-ku, Fukuoka, Fukuoka, 819-0395, Japan\\
Email: kanji@mech.kyushu-u.ac.jp}
}

\maketitle
% \IEEEpeerreviewmaketitle
\pagestyle{empty}
\thispagestyle{empty}

\begin{abstract}
Vehicle relocation is the problem 
in which a mobile robot has to estimate 
the self-position with respect to an a priori map of landmarks using the perception and the motion measurements without using any knowledge of the initial self-position.
Recently, RANdom SAmple Consensus (RANSAC), a robust multi-hypothesis estimator, has been successfully applied to offline relocation in static environments. 
On the other hand, online relocation in dynamic environments is still a difficult problem, 
for available computation time is always limited,
and for measurement include many outliers.
To realize real time algorithm for such an online process,
we have developed an incremental version of RANSAC algorithm by extending an efficient preemption RANSAC scheme. 
This novel scheme named incremental RANSAC
is able to find inlier hypotheses of self-positions 
out of large number of outlier hypotheses contaminated 
by outlier measurements.
\end{abstract}

\section{Introduction}\label{sec:intro}

For safe and efficient navigation, it is crucial for a mobile robot to estimate the self-position with respect to an a priori given map of point landmarks
called global map
using the perception and the motion measurements
without using any knowledge of the initial self-position. This problem is called vehicle relocation (also "global localization", or "kidnapped robot problem"), and has received considerable attention over the last decades \cite{Thrun01}\cite{Neira03a}. 
A relocation problem can be classified into {\it offline} or {\it online}, according to whether the localization takes place {\it after} or {\it during} the robot navigation. Especially, online relocation is more challenging due to the limitation on the available computation time. Previous studies on online relocation have focused on static environments \cite{Thrun01}\cite{Lina05a} or small environments \cite{Fox99c}\cite{wolf04}, and little study has been done on relocation scalable to large dynamic environments.

RANdom SAmple Consensus (RANSAC) \cite{Martin81a} is one of most effective algorithms that have been successfully applied to {\it offline} relocation in large-scale  static environments \cite{Neira03a}\cite{Yuen05a}. 
In the context of relocation, 
RANSAC 
acts as a robust map-matching algorithm,
estimating the self-position by matching between 
the global map of landmarks
and 
a local map of observed features.
The algorithm randomly generates a set of self-position hypotheses by matching minimal sets of features and landmarks, and then scores each hypothesis by counting the number of other matches. Despite its simplicity, RANSAC performs very well in the presence of outlier features. In the famous study carried out by Neira, Tard\acu{o}s and Castellanos \cite{Neira03a}, the effectiveness of RANSAC-based relocation over conventional algorithms was demonstrated. However, the original RANSAC is essentially {\it offline} algorithm and is not directly applicable to the online relocation,
where features incrementally arrive and available time is always limited.  Moreover, its computational complexity
depends strongly on the number of features and outliers.

This paper focuses on the problem of RANSAC-based {\it online} relocation in large-scale dynamic environments. 
In such an online process, 
the robot can no longer score all the hypotheses by the full set of features due to the limitation on the computation time,
instead it can only score 
a small set of hypotheses by a small set of features.
So,
the robot is required 
to plan the order in which hypotheses and features are scored,
so as to avoid excessive scoring of useless hypotheses or features,
which are contaminated by outliers inherent in large or dynamic environments.
To solve this problem,
we propose to utilize an efficient RANSAC scheme, 
called preemption scheme,
which was 
originally proposed for
computer vision applications by Nister \cite{Nister03a}.
The only difference of this preemptive RANSAC from the standard one is that the scoring order is planned to achieve high efficiency in the limited time. Importantly, 
this technique puts no constraint on the hypothesis generation, 
e.g. avoiding any guided hypothesis generation. 
We have developed an incremental version of the preemptive RANSAC, 
named incremental RANSAC,
which can handle incrementally arriving features
while requiring a constant time per viewpoint.
We apply this novel RANSAC scheme
for matching the global map and a local map that is incrementally updated  in a constant time by utilizing Sparse Extended Information Filter (SEIF). As a result, our real-time algorithm worked robustly in large dynamic environments even where more than $50\%$ of the landmarks have been modified.

\section{Related Works}

Previous techniques for online localization can be classified into two categories, according to whether the initial self-position is known or not. If the initial self-position is known, the localization problem is equivalent to position tracking, and traditional techniques 
such as Kalman Filtering  \cite{wolf04}\cite{Smith91a} are applicable. If the initial self-position is unknown, the full relocation problem needs to be solved.

Markov Localization and Monte Carlo Localization \cite{Thrun01} are two popular algorithms for online relocation. They generate a number of self-position hypotheses covering all possible positions and score the likelihood of every hypothesis based on consistency between features and landmarks. Although they are reliable in relatively small environments \cite{Lina05a}, they are not scalable to large environments since the number of required hypotheses is linear to the environment size. 

There are also some offline algorithms that are scalable to large environments. 
They aim to estimate the self-position by matching 
between the global landmark map 
and a local feature map. The essence of these algorithms is to generate a small set of good initial hypotheses 
by matching minimal set of features and landmarks \cite{Neira03a}\cite{Yuen05a}. 
As briefly described in section \ref{sec:intro}, RANSAC is one of such algorithms.
Their computational cost depends not on the environment size but on the number of features, therefore they are efficient especially in sparse environments \cite{Lina05a}. However, even these algorithms are not directly applicable to online relocation, where available computation time is always limited and typically constant. Moreover, larger number of features would be required  in the case of dynamic environments
since there are usually many outlier features. This makes it difficult even to apply some pre-computed lookup tables that could accelerate the map matching \cite{Fox99c}.

\section{Background}

\subsection{Problem}\label{sec:problem}

\begin{figure}[t]
\begin{center}
\FIG{8}{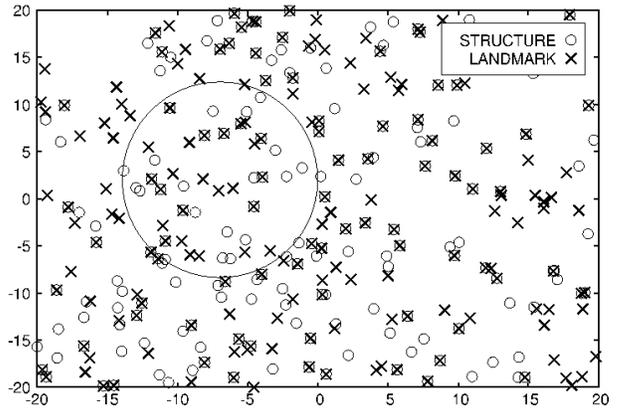}{}
\caption{A small $400[m^2]$ area in an environment
(change ratio: $50\%$).}\label{fig:motivation}
\end{center}
\end{figure}

The goal of relocation
is to estimate
the self-position of the robot 
with respect to a priori given map
by using motion and perception sensors
without using any a priori knowledge 
of the initial self-position.
Online relocation is a process to
incrementally update the belief of the self-position
every time new motion and perception measurements arrive. 
Following the SLAM
literature \cite{Thrun02},
the a priori map
is called 
{\it global map}
and represented by a set of 2D point landmarks.
Each landmark is a cartesian pair.
There is no distinctive landmark
that is distinguishable from another landmark.
The motion sensor acquires
the ego-motion of the robot on the floor plane. 
The perception sensor 
acquires relative position to the surrounding landmarks. 
A popular example of the motion and the perception sensors
is omni- laser scanner and wheel encoder. 
Fig. \ref{fig:motivation} shows the area observed by the perception sensor at a certain viewpoint as a circular region.

Features observed at more than one viewpoint 
are often required to uniquely determine the self-position.
In such a case,
it is useful
to construct a {\it local map} of features
from scratch 
through a SLAM process 
during the robot navigation.
When the local map grows sufficiently informative,
the self-position
will be uniquely determined 
by matching the global and the local maps.

We will utilize SEIF \cite{Thrun04a},
a constant time SLAM filter, 
to update the local map as well as the self-position with respect to the local map.
Instead of traditional representation in Kalman Filter using a mean vector and a covariance matrix, SEIF represents the current estimate of the system state
(i.e. self-position and map)
by so-called information matrix and information vector. Information matrix is the inverse of the covariance matrix, and naturally sparse i.e. it tends to have much less strong constraints than the covariance matrix. In SEIF, the sparseness of the information matrix is enhanced by a process called sparsification that eliminates weak robot-landmark constraints. As a result, the perception and the motion update can be done in a constant time. The local map can be also recovered from the information matrix and the information vector in a constant time by a process called amortized map recovery.
For more details of SEIF, 
see \cite{Thrun04a}.

In large dynamic environments, there are mainly two sources of outlier features. First, the robot is not necessarily located in the mapped area,
the area covered by the global map. 
In addition, the robot 
does not know whether it is located in mapped or unmapped area. Therefore, while it navigates in unmapped area, it obtains only outlier measurements. Second, some landmarks may have been modified due to environment changes. 
Fig. \ref{fig:motivation} illustrates an example of such environment changes.
In this figure,
'STRUCTURE' and 'LANDMARK' points respectively indicate 
landmark locations after and before the environment changes.
Landmarks can be removed, added or even moved from one place to another. 
Such dummy (changed) landmarks 
tend to increase outlier measurements. Moreover, removed landmarks will cause even reduction of inlier measurements. Therefore, we view this second outlier source is more critical. All of above mentioned outliers are not simply distinguishable from inliers.

\subsection{Probabilistic Formulation}

The relocation problem is formulated in a probabilistic framework.
In the following, 
we denote by $v^k$ a sequence $(v_1, \cdots, v_k)$.
Let $M$ denote the a priori given global map. $Z_t$ denote measurement arrives at time $t$. $Z_t$ can be a motion or perception measurement. Without loss of generality, we assume each perception measurement corresponds to exactly one landmark in the environment. 
Let $L_t$ denote the set of parameters estimated by SEIF,
i.e.
the local map and the self-position with respect to the local map.
Since the global map and the local map are represented in different coordinate systems, the coordinate transformation (rotation and translation)
from the local map to the global map 
also needs to be estimated.
Let $\psi$ denote such a transformation, which we call 
{\it global position}.
Then,
the system state to estimate is represented by $G_t=(\psi, L_t)$.
Based on the above terminology, the online relocation is formulated as the problem of incrementally 
updating 
the belief from $P(G_{t-1}|Z^{t-1})$ 
to $P(G_t|Z^t)$ 
every time new measurement $Z_t$ arrives.
This probability density 
$P(G_t|Z^t)$
can be decomposed in the form
\eq{
P(G_t|Z^t) = P(L_t|Z^t) P(\psi|L_t, Z^t),
\label{eq:factor}
}
then
both the terms
$P(L_t|Z^t)$
and
$P(\psi|L_t, Z^t)$
can be efficiently updated
in the following manner.
The first term
$P(L_t|Z^t)$
is the probability density
that can be estimated
from
$P(L_{t-1}|Z^{t-1})$
in a constant time 
by SEIF 
as discussed in section \ref{sec:problem}.
The second term
$P(\psi|L_t, Z^t)$
can be approximated by $P(\psi|L_t)$,
since 
$L_t$ can be viewed as
a summary of the history $Z^t$ of observations.
We view the problem of estimating
$P(\psi|L_t)$ as a kind of map matching problems, to search a best transformation $\psi$ by which the local map in $L_t$ and the global map $M$ maximally overlap.

\begin{figure}[t]
\begin{center}
\FIG{8.5}{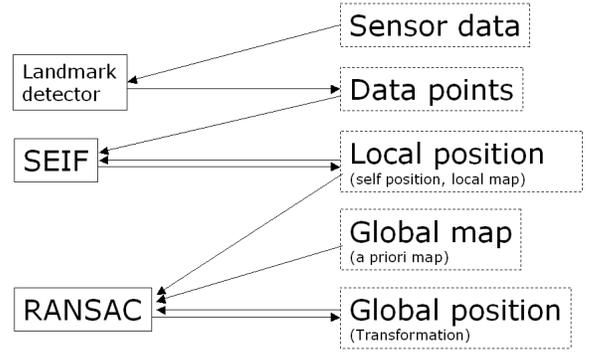}{(a) System architecture.}\label{fig:archi}
\vspace*{8mm}
\FIGR{8}{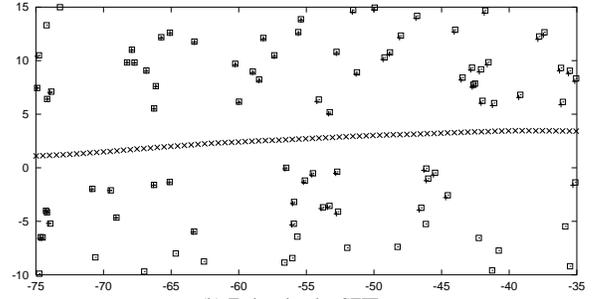}{(b) Estimation by SEIF.}\label{fig:local}
\vspace*{8mm}
\FIG{8}{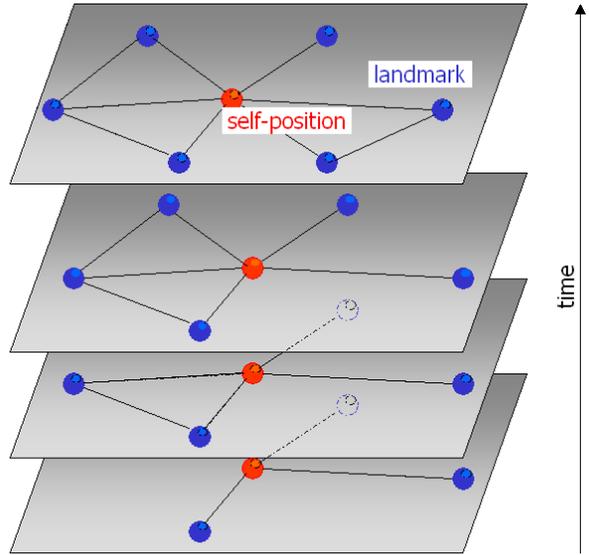}{(c) Gradually changing map.}\label{fig:gradually}
\caption{System overview.}\label{fig:overview}
\end{center}
\end{figure}

\subsection{System Overview}\label{sec:overview}

Fig. \ref{fig:overview}(a) illustrates the system architecture.
The landmark detector 
tracks and detects moving objects
based on relative speed of surrounding objects, 
and then outputs features that are likely to correspond to static landmarks.
Using such output features
as well as the motion measurements,
the SEIF filter updates the probability density 
from $P(L_{t-1}|Z^{t-1})$ to $P(L_t|Z^t)$. 
Fig. \ref{fig:overview}(b)
illustrates an example of features ('+')
estimated by SEIF,
as well as 
the true robot trajectory ('$\times$') and the true landmark locations ('$\Box$').
RANSAC module 
updates the belief of the global position $\psi$ 
by matching 
the global map and the local map.
Note that 
this matching problem 
gradually changes over time
as the local map is incrementally updated.
As shown in the simple example in Fig. \ref{fig:overview}(c),
the local map is updated
every time new features arrive or 
old features are updated by SEIF.
We will present a constant-time RANSAC algorithm
suitable for such incrementally changing
matching problems in section \ref{sec:proposed}.

\section{Incremental RANSAC}\label{sec:proposed}

\subsection{RANSAC-based Map-matching}\label{sec:map-matching}

RANSAC is a robust multi-hypothesis estimator in the presence of many outlier features. In the context of relocation, a feature $o$ corresponds to one feature in the local map while a hypothesis $h$ corresponds to one hypothesis of global position $\psi$, a transformation from the local to the global coordinate. RANSAC is essentially an offline algorithm, 
for it assumes all the features are a priori given.

The algorithm is summarized as follows. In the initialization stage, the system randomly permutes the features. Then, it iterates the following steps until the time budget is exhausted.
\begin{enumerate}
\item
Generate a hypothesis $h$ 
from randomly selected $k$ features
by matching the $k$ features with landmarks.
\item
Initialize score $s_h$ of the hypotheses $h$ to $0$.
\item
For each feature $o$,
check if
$(o, h)$
is an inlier pair,
and if so increment
$s_h$.
Here,
$(o, h)$ is judged as inlier iff 
the feature $o$ transformed to the global coordinate by $h$
is located within a neighborhood 
of some landmark.
\end{enumerate}
When the process finishes,
a hypothesis with highest score 
is output as the best transformation $\psi$.
Usually,
the computational cost
of the step 1 or 2
is negligible compared to the cost of step 3,
therefore the total cost is approximately linear to the number of 
feature-hypothesis pairs.

The parameter $k$ should be carefully determined
taking into account
the tradeoff between 
efficiency and reliability.
In step 1,
at least two features 
are required to uniquely determine a hypothesis 
i.e. the global position $\psi$,
therefore $k\ge 2$.
As $k$ increases,
the easier it is to reduce outlier hypotheses by using the 
($k-2$) redundant features, therefore it becomes easier to increase the efficiency.
On the other hand,
as $k$ decreases,
the higher the probability of 
at least one hypothesis being inlier, 
therefore the reliability increases.
We set $k=3$,
and randomly select 
such $k$ features 
that are covisible from at least one viewpoint.
This is similar in concept with
the 'locality' suggested in \cite{Neira03a}.
We found that this procedure generates almost constant number of new hypotheses every time a new feature arrives.

\subsection{Preemptive RANSAC}\label{sec:preemption}

\begin{figure}[t]
\begin{center}
\FIGR{9}{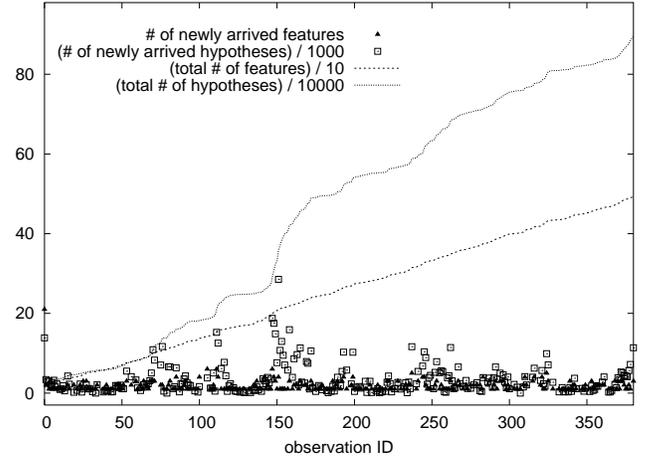}{}
\caption{The number of features and hypotheses.}\label{fig:oh}
\end{center}
\end{figure}

Even with the above procedure,
the number of hypotheses 
is linear to the number of observations.
Fig. \ref{fig:oh}
shows how 
the number of features and hypotheses grows as the robot navigates.
Obviously,
it is difficult for the original RANSAC
to achieve constant-time online relocation.

It can be viewed that the original RANSAC scores
all the feature-hypothesis pairs $\{(o, h)\}$ excessively. 
Although such an algorithm is reliable, 
its computational cost is at least the square order of the number of observations. 
Instead of such an excessive scoring,
the preemption scheme proposed by 
Nister \cite{Nister03a} 
plans {\it scoring order} 
in which pairs are scored
so as to avoid scoring useless pairs.
In the following, we briefly summarize the scheme. 

A preemption scheme $\Omega$ is composed of an {\it order rule} $f_o$ and a {\it preference rule} $f_p$: 
\eq{
\Omega=(f_o, f_p).
} 
An order rule determines the next pair 
$x_j=(o_j, h_j)$
to be scored given all the previous scoring results:
\eq{
x_j=f_o(x^{j-1}, s^{j-1}). 
}
Here, $s_j$ represents the score of the hypothesis $h_j$.
A preference rule selects the most preferred (best) hypothesis given all the
scoring results: 
\eq{
h^{best}_j=f_p(x^j, s^j). 
}
A preemption scheme is called {\it depth-first} 
if the order obeys
\eq{
h_{j_1} \le h_{j_2}
~~~~\forall (j_1, j_2) : j_1\le j_2,
}
or called {\it breadth-first} 
if the order obeys 
\eq{
o_{j_1} \le o_{j_2}
~~~~\forall (j_1, j_2) : j_1\le j_2,
}
or called {\it hybrid} otherwise.
In the next section, 
this efficient offline scheme 
will be extended for our 
{\it online} relocation problem.

\subsection{Incremental Scheme}

To deal with incrementally arriving features, 
we have developed a novel RANSAC scheme,
named incremental RANSAC,
which combines
the advantages of 
the RANSAC-based map-macthing (section \ref{sec:map-matching})
and
the preemptive RANSAC (section \ref{sec:preemption}).
This novel scheme has the following two characteristics:
\begin{enumerate}
\item
Incremental estimation:
at every viewpoint,
feature-hypothesis pairs are scored
conditioned on the scoring results at the previous viewpoints.
\item
Real-time estimation:
at every viewpoint,
only a constant number of feature-hypothesis pairs 
are selected 
to be scored within the limited computation time.
\end{enumerate}

\begin{figure}[t]
\begin{center}
\hspace*{-2.3mm}\FIG{8}{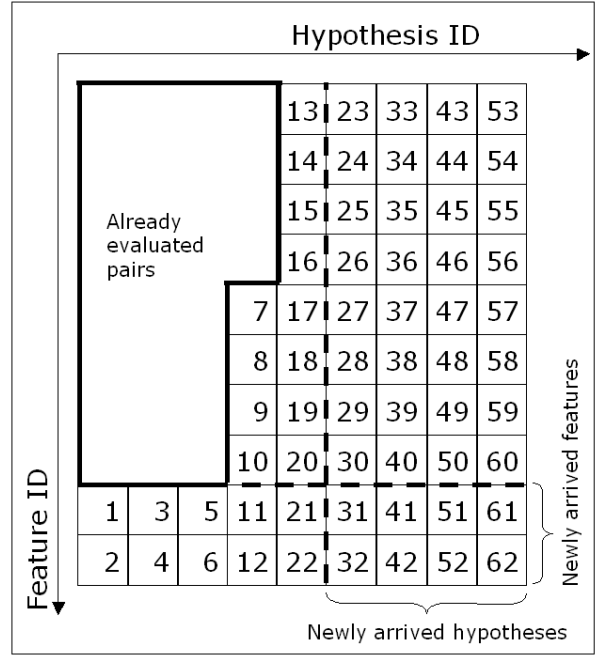}{(a) Depth-first order rule.}\\
\FIG{8}{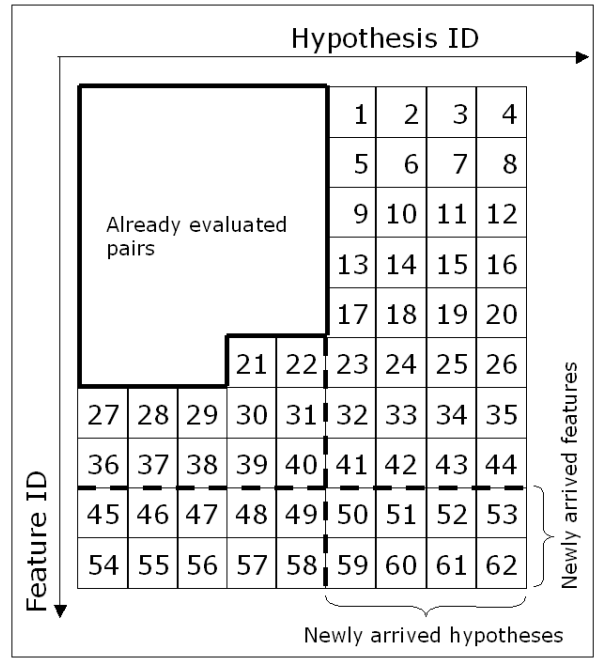}{(b) Breadth-first order rule.}
\caption{Two examples of the planned order.}\label{fig:order}
\end{center}
\end{figure}

The incremental scheme is summarized as follows.
For incremental estimation,
the scheme maintains two kinds of lists called {\it feature list} $S_o$ and {\it hypothesis list} $S_h$, which respectively record already arrived but not yet scored features and hypotheses. 
In the initialization stage at $t=1$, 
the robot sets the lists $S_o$, $S_h$ to null. 
At every viewpoint, 
firstly, the robot updates the lists $S_o$ and $S_h$ 
in the following steps,
which is essential to realize the incremental estimation.
\begin{enumerate}
\item[1-1)]
Add newly arrived features $S_o'$ to $S_o$, 
then randomly permute $S_o$.
\item[1-2)]
Generate new hypotheses $S_h'$.
Here, 
each hypothesis should be generated from randomly selected $k$ features, 
at least one of which is a newly arrived feature.
\item[1-3)]
Add $S_h'$ to $S_h$, then randomly permute $S_h$.
\end{enumerate}
Secondly, the robot iterates the following steps for $N_p$ times,
which is essential to realize the real-time estimation.
\begin{enumerate}
\item[2-1)]
Determine the next pair $(o,h)$ by the order rule.
\item[2-2)]
If $(o,h)$ is an inlier pair, increment $s_h$.
\item[2-3)]
If $o$ is a member of $S_o$, then eliminate $o$ from $S_o$.
\item[2-4)]
If $h$ is a member of $S_h$, then eliminate $h$ from $S_h$.
\end{enumerate}
Fig. \ref{fig:order} (a) and (b) respectively show examples of the orders determined by the depth-first and the breadth-first order rules in the incremental RANSAC scheme. As can be seen from the figures, incremental RANSAC is essentially a hybrid scheme. 

Generally, there is no reason to score a pair twice. 
Some simple order rules, including the depth-first and the breadth-first rules, may perfectly avoid such loss.
However,
more complex order rules 
can avoid such loss 
only by memorizing 
all of the already-scored pairs.
Unfortunately, this requires memory space proportional to the square of the number of observations.
To solve this problem,
our method clears the memory 
every time 
the robot reaches a new viewpoint.
Although this does not guarantee that no loss will occur,
this requires only constant size memory space.

\subsection{Preference Rule}

The most popular preference rule 
for the original RANSAC 
is to select a hypothesis with the highest score. However, in our case, the score does not directly reflect the likelihood of a hypothesis being inlier, since the probability of scoring each hypothesis is not equal. 
We found that the above preference rule is often unreliable in the case of incremental RANSAC. To deal with the problem, the preference rule must be given in a normalized form:
\begin{eqnarray}
f_p & = & arg~\max_h~r_h; \\
r_h & = & s_h / q_h.
\end{eqnarray}
Here, 
$q_h$
is the total number of the hypothesis $h$ being scored.
So,
$r_h$
represents the ratio of the hypothesis $h$ being scored as inlier.

\subsection{Order Rule}

Both of the depth-first and the breadth-first schemes have some appealing properties. 
\begin{itemize}
\item
A depth-first scheme tends to score small set of hypotheses by full set of features. 
Most of existing techniques for RANSAC-based relocation \cite{Neira03a}\cite{Yuen05a}
can be classified into depth-first scheme. 
Joint Compatibility Test \cite{Neira01a} is a technique to accurately discriminate between similar hypotheses using many features. 
\item
A breadth-first scheme tends to score full set of hypotheses by a small set of features. 
A breadth-first scheme has been successfully applied to a practical real-time application of Structure From Motion (SFM) \cite{Nister03a} where many hypotheses are contaminated by outlier features. 
\end{itemize}
Cleary,
they also have some limitations.
\begin{itemize}
\item
The depth-first scheme is effective 
only if the ratio of inlier hypotheses is sufficiently high \cite{Martin81a}.
\item
The breadth-first scheme is effective only if relatively 
small number of features are required to distinguish inlier hypotheses from outlier ones \cite{Nister03a}. 
\end{itemize}
Unfortunately, these are not the case in our problem. 
In large and dynamic environments, the ratio of inlier hypotheses is low, and many hypotheses are not easily distinguishable.

To solve the problems,
we propose to use some hybrid order rule 
which is based on the following criteria.
\begin{enumerate}
\item 
Diversification:
score as many hypotheses as possible. 
\item
Intensification:
score preferred hypotheses by using 
as many features as possible.
\end{enumerate}
The criteria
1) and 2)
are 
respectively 
similar in concept with the breadth-first and the depth-first schemes.
In general,
there is a tradeoff
between
the diversification and the intensification
of an optimization algorithm.

To implement the above idea, our algorithm classifies all the hypotheses into several groups 
\eq{
S_0, S_1, \cdots, S_{k-1}
}
of different preference levels according to the previous scoring results. 
Let $n(i)$ denote the number of hypotheses belonging to group \#i.
Our classification rule is to determine the group ID by
\eq{
i= 
\cases{
\lfloor k r_h \rfloor & $(r_h<1)$ \cr
k-1 & $(r_h=1)$
}.
}
The larger ID $i$ means higher preference level.
Based on the classification results, 
the algorithm randomly selects
\eq{
n'(i) = \lceil \alpha_o n(i) f_w(i) \rceil
}
hypotheses 
in total
from each group $S_i$.
Here, 
$f_w(i)$
is a weighting function called {\it preemption function}, which prioritizes different groups based on their relative importance.
$\alpha_o$ is a normalizing coefficient
that makes 
\eq{
N_p = \sum_i n'(i).
}
For each selected hypothesis,
the algorithm generates a pair 
composed of the hypothesis
and a randomly selected feature.
In this way, 
it generates $N_p$ pairs in total,
which our order rule will output one by one.

\begin{figure}[t]
\begin{center}
\FIG{8.7}{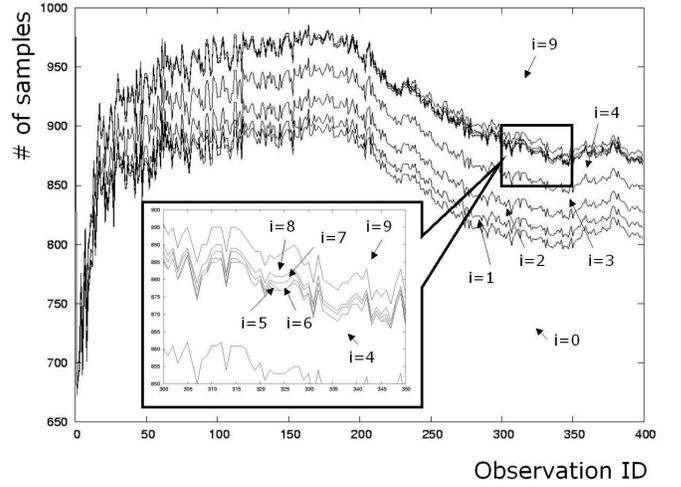}{}
\vspace*{-8mm}
\caption{Number of the samples from each preference group.}\label{fig:samples}
\end{center}
\end{figure}

We use a simple preemption function in the form:
\eq{
f_w(i)=2^{mi} .
}
In experiments
described in section \ref{sec:simulation},
we simply set 
$m=1$.
This means that 
the probability of a hypothesis in a group $i+1$ being scored
is almost twice the probability of a hypothesis in the group $i$ being scored.
Also, we will use
$k=10$,
which means that 
$f_w(i)$
ranges from $1$ to $2^9$.

Fig. \ref{fig:samples}
shows an example of 
the number of hypotheses selected from each group
for $N_p=1000$.
In this example, the number is largest for group \#0 due to large group size $n(0)$, 
and next largest for group \#9 due to high weight $f_w(9)$.
Note that 
these two results respectively correspond to
the two criteria,
diversification and intensification,
described above.

To determine the next pair,
our algorithm
determines the hypothesis
{\it before}
it randomly selects the feature.
Note that 
once it determines the hypothesis 
i.e. the transformation $\psi$,
it can uniquely determine the location
of any (already-arrived) feature
with respect to the global map.
Therefore,
there is no reason 
to select a feature located in an unmapped area.
Our algorithm can avoid selecting such useless features 
by selecting the feature 
in the following steps.
\begin{enumerate}
\item
Randomly sample a location $l$ in the mapped areas.
\item
Output such a feature 
that is nearest to $l$.
\end{enumerate}
The search process in the second step 
can be executed quite fast 
based on a space division technique using quadtree.
Such a selection strategy 
must be 
important especially when the ratio of outlier features becomes large.

\newpage

\section{Experimental Evaluation}\label{sec:simulation}

\begin{figure}[t]
\begin{center}
\FIG{8.5}{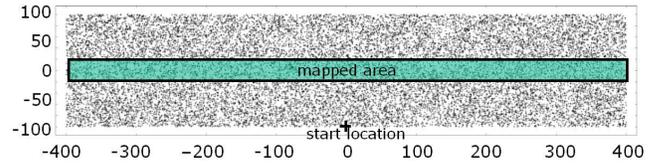}{}
\caption{Landmarks, start location, and mapped region.}\label{fig:env}
\end{center}
\end{figure}

To evaluate the robustness of the proposed method against outlier observations, we conducted a number of experiments in virtual environments. The robot is equipped with an omni- laser scanner that can observe landmarks within $10[m]$, with Gaussian noise with $0.01[m]$ standard deviation in range and $0.5[deg]$ in azimuth. The robot is also equipped with a wheel encoder that measures translation and rotation movement with Gaussian noise with $1\%$ standard deviation. Fig. \ref{fig:env} shows an example of the environments. The environment size is $800[m]\times 200[m]$. The robot is initially located at $(0, -100)$, and moves towards the goal location at $(0, 100)$. Every time it moves $0.5[m]$, it observes the surroundings by the laser scanner. 
Based on the latest measurements, it updates its belief by the SEIF as well as the incremental RANSAC. 

Every virtual environment is generated in the following manner. Mapped area is $[-400, 400]\times[-20, 20]$, 
only $1/5$ of the entire environment. $20000$ landmarks are randomly distributed in the environment. A predetermined ratio (ranging from $0\%$ to $99\%$ as described later) of landmarks have been moved due to environment changes. Such high landmark density, small mapped area, as well as environment changes cause many outlier features and hypotheses, which make our relocation problem more difficult. 

To evaluate the performance for various outlier ratio, we generated $100$ test environments of different ratio of changes, ranging from $0\%$ to $99\%$. We tested the proposed method and benchmark against two other methods, a depth-first method and a breadth-first method described in \ref{sec:proposed}. To achieve real-time processing, $N_p$ had to be set to a small number, $1000$.
The computational costs of our relocation
process per viewpoint (CPU Pentium4 $1.8[GHz]$, $256[Mbyte]$) 
are as follows.
\begin{itemize}
\item
SEIF: $350.4[ms]$ 
\begin{itemize}
\item
perception update: $11.7[ms]$
\item
motion update: $7.9[ms]$
\item
amortized map recovery: $36.9[ms]$
\item
the others (including the sparsification): $294.0[ms]$
\end{itemize}
\item
RANSAC: $308.0[ms]$ ($N_p=1000$)% 123.210000 / 400
\end{itemize}
The total computational cost is less than $1[s]$, which is sufficiently low cost for many real-time applications \cite{Thrun02}.

\begin{figure}[t]
\begin{flushright}
\hspace*{-4.5mm}\FIGR{9.5}{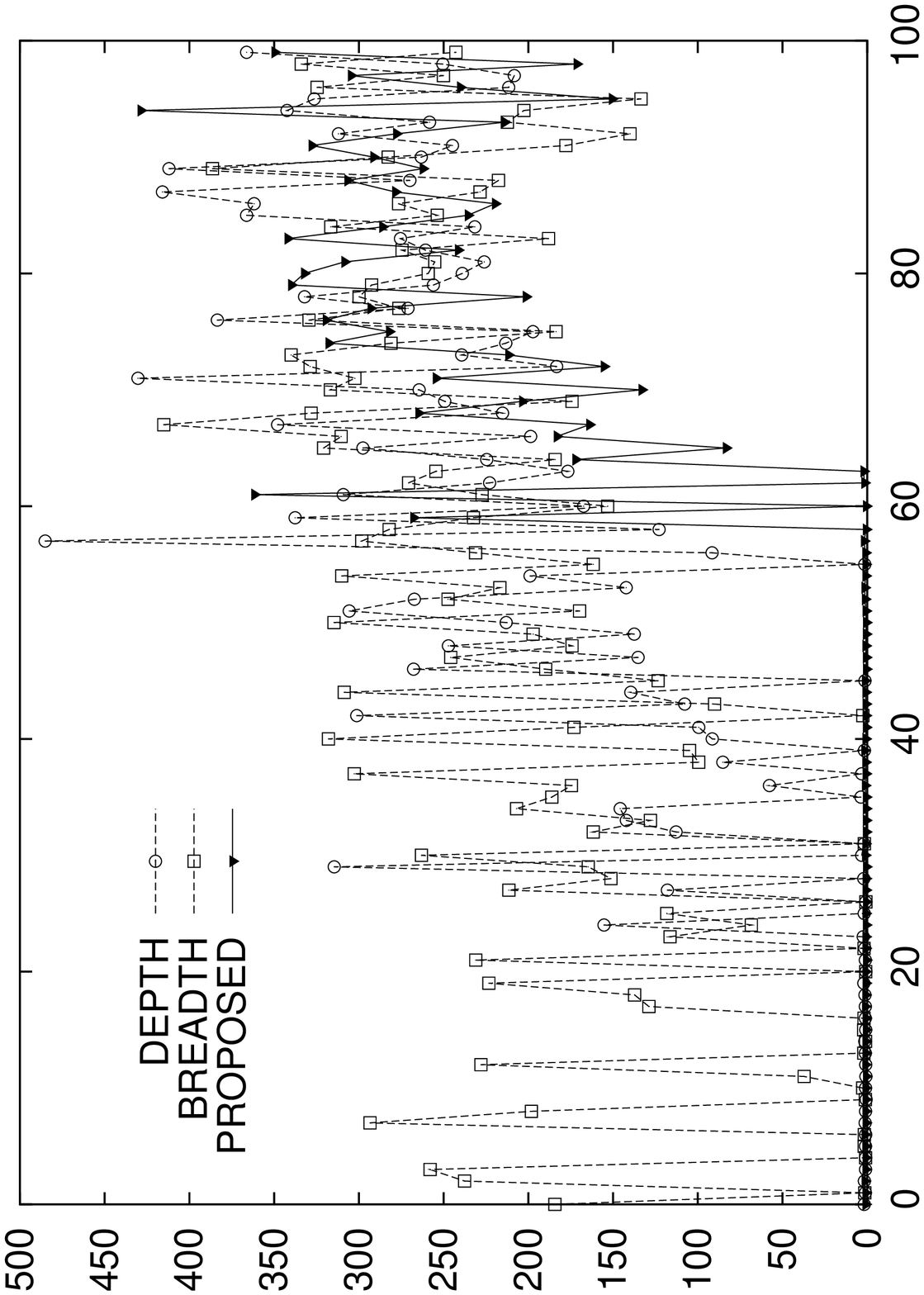}{\vspace*{-2mm}}\\
\FIGR{9.1}{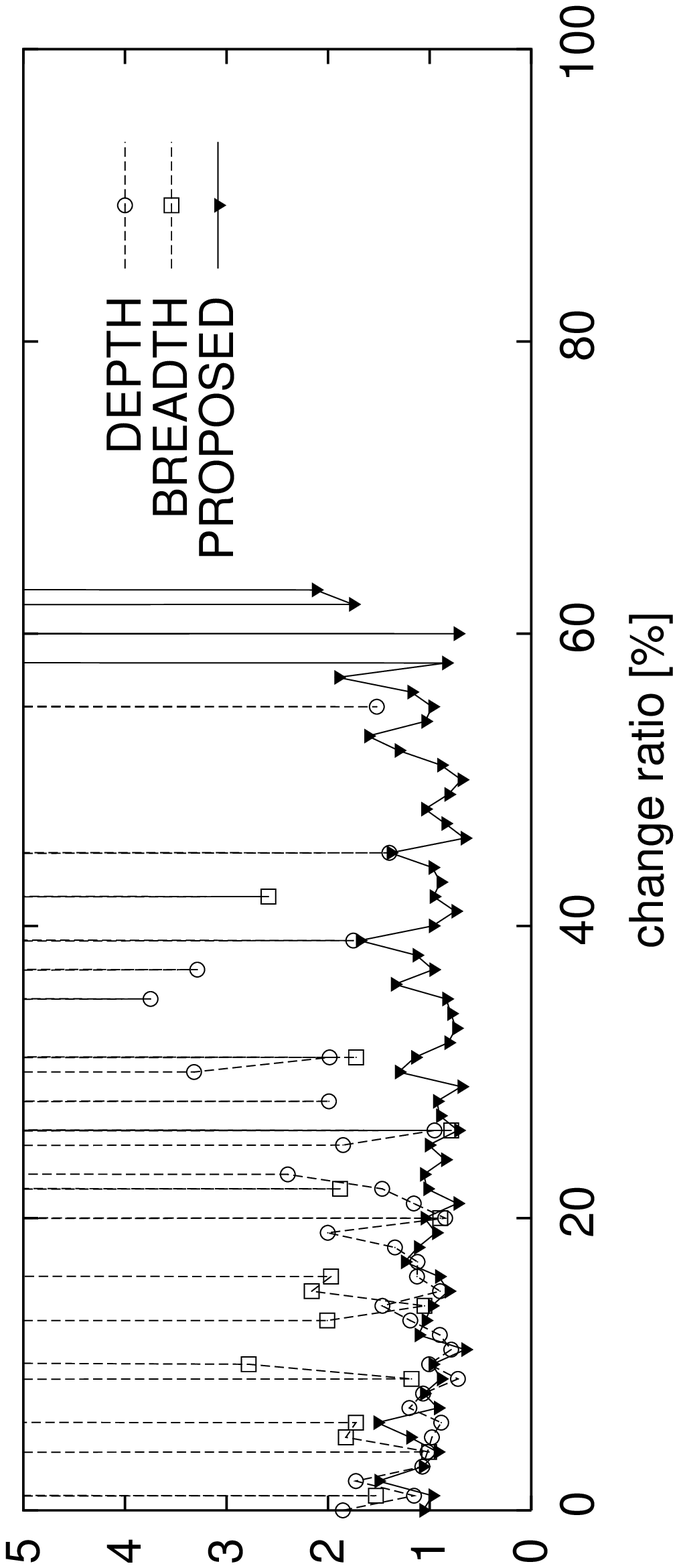}{}\hspace*{-2.6mm}
\end{flushright}
\vspace*{-5mm}
\begin{center}
\caption{Performance for different ratio of changes.}\label{fig:performance}
\end{center}
\end{figure}

Fig. \ref{fig:performance} summarizes
the results.
The horizontal axis indicates the ratio $\%$ of changed landmarks, while the vertical axis indicates the localization error $[m]$ at the (estimated) goal location. 

The depth-first scheme was successful when the order rule selects at least one inlier hypothesis. However, the probability of selecting such an inlier hypothesis in the limited computation time is inverse proportion to the number of the pairs. 
As a result, 
the method became unreliable 
if the change ratio exceeds $25\%$.

The breadth-first scheme gives the maximum probability of scoring inlier hypotheses over all possible preemption schemes. However, 
the method could not score inlier hypotheses by many features due to the limitation on the computation time. 

On the other hand, our hybrid scheme performed very well. Especially, when the change ratio is less than $58\%$, 
the method achieved low localization error, less than $2[m]$. 
Like the breadth-first scheme, the method gives a high probability of scoring inlier hypotheses. Also, like the depth-first scheme, the method scores preferred hypotheses (i.e. hypotheses in high preference groups) by many features. 
From the results, it can be concluded that the proposed incremental RANSAC gives a scheme of a real-time process for robust online relocation in large and dynamic environments.

\section{Conclusions \& Future Works}

In this paper,
the problem of online relocation in large dynamic environments was addressed. 
To make the problem computationally tractable, the problem was decomposed into a problem of local map updating, and a problem of map matching. 
The map matching problem gradually changes over time
due to the local map updating 
as well as incrementally arriving features. 
To deal with the problem, 
an efficient RANSAC scheme called preemptive RANSAC was modified, 
and an incremental version named incremental RANSAC was proposed. 
The robustness of the proposed algorithm  against outlier observations 
was demonstrated in a number of large and dynamic environments.

In the future,
we will develop our research along two lines.
First,
we plan to implement and evaluate
the proposed algorithm
on a real mobile robot.
In complex and less sparse environments,
transformation of real range data to feature 
may be
itself a difficult problem.
Some grid-based representation \cite{Hahnel03}\cite{Gutmann02} of the map
instead of the feature-based representation chosen in this paper,
may be effective to deal with such ambiguity.
In addition,
the landmark detection problem described in section \ref{sec:overview},
which was simplified in this paper,
will become more challenging task.
Second,
we plan to study 
a way to 
extend the proposed relocation techniques
for SLAM problems,
where the landmarks should be learned 
even in large and dynamic environments.

\section*{Acknowledgment}

This research was partially supported by the Japan Ministry of Education, Science, Sports and Culture, Grant in-Aid for Young Scientists (B), 17700200, 2005.

\bibliographystyle{unsrt}
\bibliography{paper}

\end{document}